\documentclass[journal]{IEEEtran}
\usepackage{graphicx}
\graphicspath{{figures/}{./}}
\usepackage{amsmath,amssymb}
\usepackage{booktabs}
\usepackage{array}
\usepackage{url}
\begin{document}
\title{HybridInfer: Thermal-Aware Reinforcement-Learning Tier Routing for On-Device, Edge, and Cloud LLM Inference}
\author{Simran~Koul%
\thanks{S. Koul is an independent researcher (e-mail: simrankoul2026@gmail.com).}%
\thanks{Data, workload, and code are released to support replication.}}
\markboth{}{Koul: HybridInfer: Thermal-Aware RL Tier Routing for LLM Inference}
\maketitle
\begin{abstract}
On-device inference with small language models is attractive on modern smartphones because it keeps
user data local, works offline, and incurs no per-query cost, so the on-device tier is the preferred
place to answer a query when it is adequate. On-device inference is thermally constrained, however, and I find the constraint is sharper than a slowdown: on a flagship Snapdragon device, sustained on-device generation destabilizes the GPU inference runtime, which crashes or silently wedges after a few consecutive queries. The failure lies in the current on-device toolchain (OpenCL kernel compilation and long-prompt prefill on the mobile GPU), recurs even when the device is thermally cool, and is worst for long generations. Multi-tier routers that spread queries across on-device, edge, and cloud models can relieve
this pressure, but existing routers are thermal-blind and are typically evaluated in simulation or on
non-mobile hardware. I present HybridInfer, a thermal-aware reinforcement-learning router for a
three-tier hierarchy (on-device Llama 3.2 3B, edge Llama 3.1 8B with retrieval, and cloud GPT-4o) that
uses the phone's thermal headroom and a query-complexity estimate as state and selects a tier by an
offline-trained Q-learning policy. Its reward trades answer quality against latency, cost, and a thermal
penalty, plus a locality bonus that credits on-device execution. I show this bonus is a precondition
for thermal-aware routing to be meaningful: without it the optimal policy offloads every query. On a real Android benchmark of 210 prompts, the learned router attains significantly higher quality than two hand-tuned heuristics (paired Wilcoxon, $p < 0.02$) at the lowest cost of any adaptive condition. The always-on-device conditions are competitive in per-query quality on the queries they can serve, but are three to six times slower than the offloading tiers and fail outright on long queries, so routing wins on latency, reliability, and coverage rather than on quality. To my knowledge this is the first use of on-device thermal
headroom to select among LLM inference tiers of differing capability on real hardware.
\end{abstract}
\IEEEpeerreviewmaketitle
\section{Introduction}
\label{sec:intro}
\IEEEPARstart{O}{n-device} inference with small language models (SLMs) has become practical on flagship
smartphones. A 3-billion-parameter model quantized to four bits fits in memory on a current
Snapdragon-class device and runs on the mobile GPU, and keeping inference on the device offers real
advantages: user data never leaves the phone, the service works offline, and there is no per-query
monetary cost. These properties make the on-device tier the preferred place to answer a query whenever
it is adequate.

On-device inference is, however, thermally constrained, and my measurements show the problem is sharper than a gradual slowdown. On a Samsung Galaxy S25+ (Snapdragon 8 Elite, Adreno 830), sustained on-device generation repeatedly destabilized the GPU inference runtime, with the application crashing or silently wedging after a small number of consecutive on-device queries. Inspecting the device crash logs, the failures are in the on-device GPU inference stack, the MLC-LLM runtime and the Adreno OpenCL driver, rather than in application logic: OpenCL kernel-compilation errors, intermittent driver-library load failures, and prefill-phase failures on long prompts. Because MLC-LLM is among the most mature on-device LLM runtimes available, this reflects the current state of on-device deployment tooling rather than a fringe library. The failures recur even when the device is thermally cool, so the unreliability is only partly thermal: on-device inference with the current toolchain is not only slower under load, it is less reliable under load. Existing measurement studies of mobile LLM
inference characterize throughput and thermal throttling as a fixed performance ceiling
\cite{pockets,lmmeter}, but do not use the thermal state as a control signal.

A complementary line of work routes queries across tiers of differing capability, a small on-device
model, a larger edge model, and a cloud model, to trade quality against cost and latency. These
multi-tier and cascading routers decide where to run each query based on estimated difficulty or answer
confidence \cite{frugalgpt,hybridllm,routellm}. They are effective, but they are thermal-blind: the
routing objective is quality, cost, and latency, and the physical state of the device that actually
executes the local tier does not enter the decision. They are also, in the systems that are closest to
mine architecturally \cite{consroute,eacorag}, evaluated in server simulation or on laptop and
workstation GPUs rather than on a real phone.

I present HybridInfer, a thermal-aware reinforcement-learning router for a three-tier LLM inference
hierarchy on real Android hardware. The three tiers are an on-device SLM (Llama 3.2 3B via MLC-LLM), an
edge model with retrieval augmentation (Llama 3.1 8B with dense retrieval, served on a local host), and
a cloud model (GPT-4o). The router's state combines the on-device thermal headroom reported by the
Android \texttt{getThermalHeadroom} API with a query-complexity bin, and it selects a tier by a tabular
Q-learning policy trained offline. The reward trades answer quality, measured against an independent
gold model, against latency, monetary cost, and a thermal-throttle penalty, plus a locality bonus that
credits keeping inference on the device. The locality bonus is not incidental: without it, the optimal
policy degenerates to offloading every query, because the edge and cloud tiers are faster and comparable
in quality and do not heat the phone, and in that regime the thermal state is irrelevant to the
decision. The locality bonus is what makes the device tier preferable when it is thermally feasible, so
that the router keeps work local until thermal pressure makes offloading worthwhile.

This paper makes the following contributions.

\begin{enumerate}
\item \textit{Thermal-headroom tier selection.} To my knowledge, HybridInfer is the first system to
use on-device thermal headroom as the signal for selecting among LLM inference tiers of differing
capability, explicitly trading response quality for thermal sustainability, and to validate this on real
Android hardware. This extends thermal-aware reinforcement learning from the frequency-scaling axis,
where prior work adjusts DVFS settings of a single on-device model, to the model-capability axis, where
thermal relief is obtained by moving work to a different, higher-capability tier at the cost of privacy,
offline operation, and money.

\item \textit{An on-device instability finding, with mechanism.} I report direct evidence, backed by device crash logs, that sustained on-device LLM inference on a flagship mobile SoC is unstable, not merely slow. Measured one query at a time, the on-device tier is competitive in quality on short and medium prompts, but a continuous always-on-device session crashes or wedges within a few queries, and long prompts fail outright (they wedge the prefill phase). The root causes are in the GPU inference stack (OpenCL kernel compilation and prefill state), not only thermal throttling, since the failures recur when the device is cool. Because MLC-LLM is among the most mature on-device runtimes, this is best read as the current on-device toolchain not yet being robust for sustained or long-context inference. Thermal- and load-aware routing is thus a reliability and coverage mechanism, not only an efficiency mechanism.

\item \textit{A reward-design insight.} I show that a multi-tier router whose reward values only
quality, cost, and latency learns to offload unconditionally, which makes thermal state and indeed the
on-device tier itself irrelevant. Valuing on-device locality explicitly, as privacy, offline
capability, and zero marginal cost, is a precondition for thermal-aware routing to be a meaningful
objective. This observation is independent of the specific tiers or hardware.

\item \textit{An open, real-device benchmark.} I release a reproducible workload of 210 prompts with
frozen gold references, an Android measurement harness, and the full routing and evaluation pipeline,
all exercised on a real Snapdragon phone, in contrast to the simulated or non-mobile setups used by the
closest prior systems.
\end{enumerate}

My experiments show that the learned router attains significantly higher answer quality than two
hand-tuned routing heuristics at the lowest cost of any adaptive condition. I am explicit about the
scope of these results, which come from a three-replication run on an unplugged, uncooled device, and
about their gaps, which are stated in the limitations.

\section{Related Work}
\label{sec:related}

\subsection{LLM query routing and cascading}
A body of work reduces the cost of serving LLM queries by routing each query to a model of appropriate
strength. FrugalGPT \cite{frugalgpt} cascades from cheaper to stronger models and stops when an answer
looks adequate. Hybrid LLM \cite{hybridllm} trains a router that sends easy queries to a small model and
hard queries to a large one, using a difficulty estimate. RouteLLM \cite{routellm} and related systems
learn to route between a strong and a weak model to preserve quality at lower cost. HybridInfer's
complexity-based and confidence-based conditions are instances of this family and serve as baselines.
The distinction is that these systems select among cloud-hosted models or a two-model strong-or-weak
pair, optimize quality against cost or latency, and do not model the physical device that executes a
local tier.

\subsection{Multi-tier device, edge, and cloud LLM inference}
Closest to my architecture are systems that route across an on-device, edge, and cloud hierarchy.
ConsRoute \cite{consroute} performs three-tier cloud, edge, and device adaptive routing of LLM queries
with a consistency-aware reranker, and reports substantial latency and cost reductions at near-cloud
quality. EACO-RAG \cite{eacorag} places retrieval-augmented generation as a distinct edge tier within a
device, edge, and cloud gating hierarchy. I do not claim the three-tier arrangement, nor RAG as an edge
tier, as novel: these systems establish that architecture. What separates HybridInfer is the routing
signal and the deployment. ConsRoute and EACO-RAG use no thermal signal; ConsRoute's device tier is a
laptop and its edge and cloud tiers are workstation GPUs, and EACO-RAG is evaluated entirely in server
simulation. HybridInfer's routing decision is driven by the thermal state of a real smartphone, and the
on-device tier runs on that phone.

\subsection{Thermal-aware inference and scheduling}
Thermal management for inference has been studied mainly at the level of frequency scaling and task
placement on a single device. The closest prior work, EdgeEngine \cite{edgeengine}, is a thermal-aware
reinforcement-learning controller: it uses reinforcement learning with temperature as part of the state,
but it controls the DVFS frequency of CPU, GPU, and memory on one embedded device (a Jetson TX2) running
CNNs under controlled thermal conditions. Frequency scaling leaves the model, and therefore the output
quality, unchanged, and EdgeEngine's own discussion names accuracy and multidimensional constraint
spaces as future work. HybridInfer takes up exactly that axis: it extends thermal-aware
reinforcement-learning control from the frequency-configuration space, where output quality is fixed, to
the model-capability-tier space, where thermal relief is purchased with a change in the answering model
and hence in response quality. Related device-side work includes thermal-aware scheduling between the
GPU and NPU for on-device DNNs \cite{tancao} and energy- and thermal-aware adaptation of on-device LLM
inference that predicts thermal state to drive its controller \cite{enerinfer}; both keep the same model
and adjust how it runs rather than which tier answers. Temperature-aware offloading has been studied for
extended-reality workloads \cite{tao}, which weighs on-device against edge execution across power,
temperature, and energy time scales, but for rendering rather than capability-tier selection among
different LLMs. Datacenter thermal-aware schedulers such as TAPAS \cite{tapas} operate on GPUs and
virtual machines rather than a device, edge, and cloud capability hierarchy.

\subsection{Reinforcement learning for LLM routing}
Recent work casts LLM routing as a sequential decision problem solved with reinforcement learning.
Router-R1 \cite{routerr1} performs multi-round routing with a cost-aware reward, and SeqRoute
\cite{seqroute} does global budget-aware sequential routing trained with offline reinforcement learning.
HybridInfer also uses offline-trained reinforcement learning, but the state that couples its decisions
across time is different in kind. In these systems the coupling variable is a spend-down token or cost
budget, an accounting quantity. In HybridInfer the coupling variable is a physical thermal state that
the agent's own tier choices change: choosing the on-device tier heats the device and reduces future
headroom, while offloading lets it cool. The novel driver is thermal-state accumulation on real
hardware, not sequential reinforcement learning as such.

\subsection{On-device LLM measurement}
A growing set of measurement studies characterizes LLM inference on commodity mobile hardware, including
throughput and thermal-throttling behavior on commercial devices \cite{pockets} and latency profilers
for on-device LLMs built on the MLC and TVM stack \cite{lmmeter}. These studies motivate the problem by
documenting that sustained mobile inference throttles. HybridInfer builds on this by turning the thermal
signal from a measured outcome into a control input, and it adds the observation that sustained
on-device LLM inference can destabilize the runtime outright, not only reduce its throughput.

\subsection{Position}
To my knowledge, no prior system uses on-device thermal headroom as the signal to select among LLM
inference tiers of differing capability (on-device SLM, edge RAG, cloud LLM), explicitly trading
response quality for thermal sustainability, and validated on real Android and Snapdragon hardware. The
distinguishing axes are three: capability-tier selection rather than same-task offloading or frequency
tuning; an explicit quality-against-thermal tradeoff; and real-device deployment rather than simulation
or emulation.

\section{Method}
\label{sec:method}

\subsection{System and hardware}

\begin{figure*}[!t]
\centering
\includegraphics[width=0.86\textwidth]{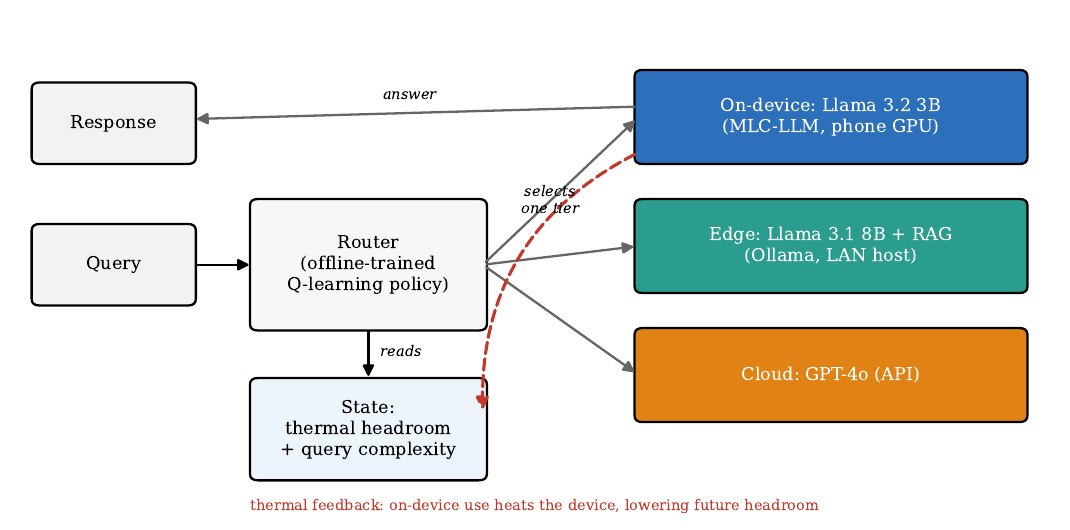}
\caption{HybridInfer routes each query to one of three capability tiers by an offline-trained
Q-learning policy whose state is the phone's thermal headroom and a query-complexity estimate.
Choosing the on-device tier heats the device and lowers future headroom (dashed feedback), which is
the physical state that couples routing decisions over time.}
\label{fig:arch}
\end{figure*}

HybridInfer routes each query to one of three inference tiers (Fig.~\ref{fig:arch}). The on-device
tier is Llama 3.2 3B
Instruct \cite{llama32} at four-bit quantization (q4f16\_0), compiled with MLC-LLM \cite{mlcllm} and
executed on the phone GPU through OpenCL. The edge tier is Llama 3.1 8B \cite{llama3} served with
Ollama \cite{ollama} on
a LAN host, augmented with retrieval (dense nomic-embed-text \cite{nomic} embeddings, cosine top-$k$
over a local corpus). The cloud tier is GPT-4o \cite{gpt4o} via the OpenAI API, streamed. The device
under test is a Samsung Galaxy S25+ (Snapdragon 8 Elite, Adreno 830 GPU). The edge host is a
LAN-connected workstation. The app records, per query, the selected tier, time-to-first-token (TTFT),
total latency, generated tokens, thermal headroom, thermal status, battery current, and the generated
text.

\subsection{Thermal signal}
The router reads Android's \texttt{getThermalHeadroom} API, a forecast in $[0, 1{+}]$ where $0$ is cool
and $1.0$ is the SEVERE-throttle threshold, together with the discrete thermal status. On the S25+ this
returns real, non-NaN values that vary with load (confirmed across all runs). The device also exposes
skin and application-processor (AP) temperatures via the thermal HAL.

\subsection{Router state, actions, and baselines}
The state is the pair (headroom bin, complexity bin), giving $5 \times 3 = 15$ states. Headroom bins are
$[0, 0.5)$, $[0.5, 0.7)$, $[0.7, 0.85)$, $[0.85, 1.0)$, $[1.0, \infty)$. The complexity bin is a
disclosed heuristic: $\mathrm{score} = \mathrm{word\_count} + 20 \cdot \mathrm{multiStepFlag}$, binned
at $30$ and $70$, where multiStepFlag fires on analytical cue words. Actions are the three tiers
($0$ on-device, $1$ edge, $2$ cloud). Seven routing conditions are evaluated: C1--C3 fix a single tier
(on-device, edge, cloud); C4 routes by complexity bin only; C5 is a confidence cascade that starts
on-device and escalates while the answer looks incomplete; C6 routes by complexity but forces offload
to edge when near throttle; and C7 is the learned Q-table, with a data-collection variant (RL\_COLLECT)
and an evaluation variant (RL\_EXPLOIT).

\subsection{Workload}
The benchmark is generated by a deterministic, seeded script that emits 60 held-out test prompts and 150
training prompts, balanced 20/20/20 and 50/50/50 across the short-factual, medium-analytical, and
long-multistep complexity tiers, and evenly spread over five technical domains. Prompts are constructed
from a curated bank of real concepts and per-tier templates. As an integrity gate, the generator
reimplements the router's complexity scoring exactly and asserts, for every emitted prompt, that its
human-readable complexity label equals the bin the router computes from the text; prompt strings are
unique and the test and training sets are disjoint. Output order is shuffled so complexity interleaves
during runs.

\subsection{Gold references and quality scoring}
Reference answers are generated once by Claude Opus 4.8 and frozen (210 references: 60 test, mean
$3322$ characters; 150 train, mean $3522$ characters). Claude is chosen as an out-of-pipeline judge: the
cloud tier is GPT-4o and the device and edge tiers are Llama, so a Claude reference does not
style-match any evaluated tier. Answer quality is scored against the frozen gold with BERTScore-F1
(roberta-large) \cite{bertscore} as the primary metric and ROUGE-L \cite{rouge} as a secondary lexical
metric.

\subsection{Reinforcement-learning router}
The Q-table is trained offline in Python, where the gold-based reward is available; the phone loads the
trained table and selects actions by argmax at evaluation time. This reference-at-training-only design
keeps the gold model out of the deployed inference path. Transitions (state, action, reward, next state)
are collected on the device and scored, then tabular Q-learning is run offline. The reward is
\begin{equation}
r = w_q q - w_l \ell - w_c c - w_t p + b_{\mathrm{local}} \, \mathbb{1}[a = \text{on-device}],
\label{eq:reward}
\end{equation}
where $q$ is the BERTScore-F1, $\ell$ and $c$ are normalized latency and cost, $p$ is a penalty rising
toward the SEVERE headroom threshold, and $b_{\mathrm{local}}$ is an on-device locality bonus. The
locality bonus credits keeping inference on the device (data locality and privacy, offline capability,
zero marginal cost); without it, edge and cloud dominate on-device in every state (they are faster,
comparable in quality, and do not heat the phone), so the optimal policy degenerates to always offload
and thermal state becomes irrelevant to the decision. The bonus makes on-device the preferred tier when
thermally feasible, while the throttle penalty overrides it when the device is hot.

\subsection{Experimental protocol and the cooled-run extension}
\label{sec:protocol}
The protocol for a fully completing, paper-grade run on a cooled device is as follows.
\begin{enumerate}
\item Cool the device to AP core temperature, not just skin, and keep it plugged (energy is then
measured in a separate short unplugged pass, or reported with the charging caveat).
\item Collect training transitions with systematic tier coverage: static C1, C2, C3 and the C6 heuristic
over the training prompts, plus deliberate heat-then-offload passes (heat with a short C1 burst, then
immediately run C2 and C3 while hot) to fill hot edge and cloud states.
\item Score training transitions against the frozen gold, train the Q-table offline, and push it.
\item Evaluate all seven conditions over the full 60 test prompts, three replications each, with
cooldowns between conditions and explicit heat-up runs so C7's offload-under-throttle behavior is
measured under real thermal stress.
\item Score and aggregate: per-condition means with 95\% confidence intervals, throttle-violation rates,
and Wilcoxon signed-rank tests of C7 against the heuristic baselines.
\end{enumerate}
The run reported below follows this protocol on an unplugged device. Because a continuous always-on-device session crashes within a few queries, the two on-device-heavy conditions (C1 and C5) cannot be run as a single sustained pass. I therefore measure them one query per fresh process launch: each test prompt is issued as the first query of a freshly started app, with a short cooldown between launches so the system-on-chip returns to a comparable temperature. This isolates per-query on-device behavior from the sustained-load crash. On-device reliably serves short and medium prompts this way, but long prompts still fail (the prefill wedges), so C1 and C5 are reported over the 40 short and medium prompts, with the long prompts summarized as a wedge rate. All C1 and C5 data is collected unplugged and verified discharging.

\section{Results}
\label{sec:results}

\subsection{Per-tier behavior}
On-device Llama 3.2 3B decodes at roughly 14 to 24 tokens/s on the Adreno 830 when the device is cool, falling toward 4 to 13 tokens/s as it throttles under sustained load (a cold first query also pays a one-time $\approx 4.8$~s TTFT for OpenCL kernel compilation, and because each single-shot launch is a fresh process it pays this each time). Measured throughput increases across tiers: on-device $\approx 14$ to $20$ tokens/s when cool, edge (8B plus retrieval) $\approx 18$ tokens/s, cloud GPT-4o $\approx 90$ to $97$ tokens/s with $\approx 0.5$~s TTFT.
This ordering is the structural motivation for routing.

\subsection{Sustained and long-context on-device inference is unstable}
\label{sec:instability}

\begin{figure}[!t]
\centering
\includegraphics[width=\columnwidth]{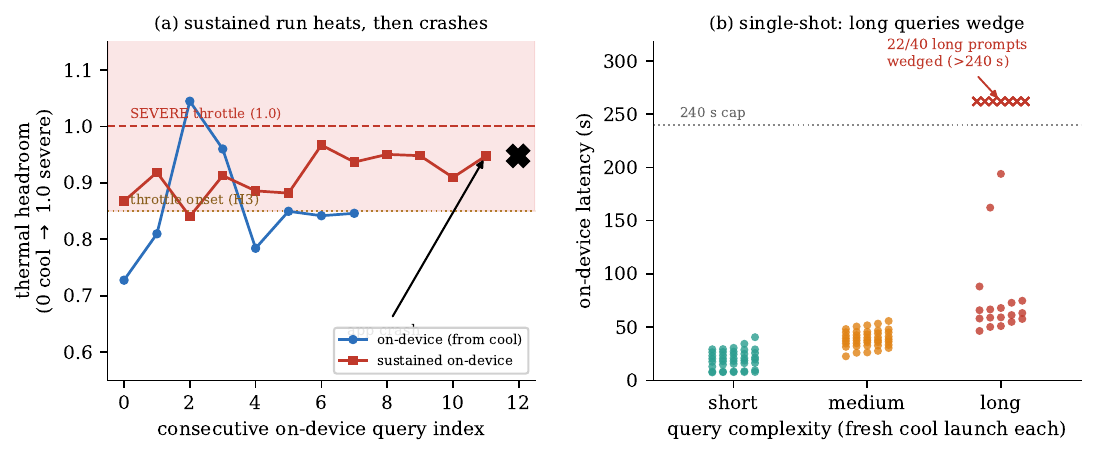}
\caption{(a) Thermal headroom over consecutive on-device queries in a continuous run: on-device inference drives headroom into the SEVERE-throttle zone (shaded) and the application crashes after 12 consecutive queries (marker); across runs it crashed at 4 to 12 queries. (b) Single-shot on-device latency by query complexity, each query issued as a fresh cool launch: short and medium prompts complete (roughly 8 to 55~s), while 22 of 40 long prompts wedge past a 240~s cap (crosses). Because each single-shot query starts from a cool state, the long-prompt wedges are not a thermal-accumulation effect.}
\label{fig:thermal}
\end{figure}

A distinct finding: under sustained on-device inference the device does not merely slow down, it becomes unstable (Fig.~\ref{fig:thermal}). In a continuous run the app crashed or was killed after roughly 4 to 12 consecutive on-device queries. The device crash logs identify the mechanism, and it is in the on-device GPU inference stack (the MLC-LLM runtime and the Adreno OpenCL driver) rather than in the application: OpenCL kernel-compilation failures, intermittent driver-library load failures, and prefill-phase failures on long prompts.\footnote{The exact runtime error signatures are included in the released crash logs.} Two observations sharpen the finding. First, it is only partly thermal: the same failures occur when the device is cool (thermal headroom near 0.6, no skin throttling), so temperature is an aggravating stressor, not the sole cause. Second, the failure has two forms, a hard crash (the runtime throws and the process dies) and a silent wedge (a long generation hangs in prefill with the process alive but producing no output). Measured one query per fresh launch, the on-device tier serves short and medium prompts reliably, but long prompts wedge: of 20 long prompts each, 9 (C1) and 13 (C5) failed to complete within a 240~s cap (22 of 40 pooled, 55\%), and the generations that did complete on-device ran for 51 to 194~s. The skin sensor recovers within one to two minutes of idle, but the application-processor core stays hot far longer. The conditions that offload once hot are the stable ones: C6 completes every replication, and the trained C7 completes the majority of its queries because it offloads most of them. Because MLC-LLM is among the most mature on-device LLM runtimes, this is best read as the current on-device toolchain not yet being robust for sustained or long-context inference, rather than a fixed hardware limit; it is nonetheless direct evidence that an always-on-device deployment on today's stack carries a stability and coverage cost, not only a latency cost.

\subsection{The learned policy is thermal- and complexity-aware}

\begin{figure}[!t]
\centering
\includegraphics[width=0.74\columnwidth]{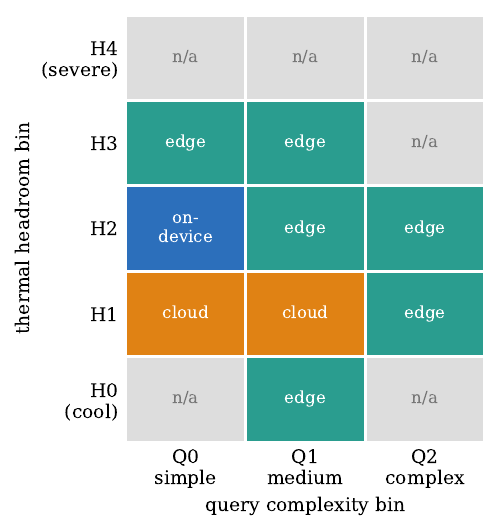}
\caption{Greedy routing policy of the trained Q-table (\texttt{trained\_128tx}) over the 15 states
(thermal-headroom bin by query-complexity bin). The learned policy keeps a simple query on-device in a
warm-but-feasible state and offloads to edge in hot states; ``n/a'' marks states not visited during
collection.}
\label{fig:policy}
\end{figure}

Trained on 128 collected transitions (on-device 11, edge 64, cloud 53) with the locality bonus, the
Q-table's greedy policy is state-dependent rather than uniform (Fig.~\ref{fig:policy}): it offloads to
edge in hot states
(headroom bin H3), keeps a simple query on-device in a cooler state where that is thermally feasible, and
routes complex queries to edge or cloud for quality. Trained on the same data without the locality
bonus, the policy collapses to always offload, confirming that the reward design, not the learning, is
what makes thermal-aware routing meaningful.

\subsection{Condition comparison}
\label{sec:comparison}
Held-out test prompts, device unplugged (clean-energy). C2, C3, C4, C6, and C7 use three replications over all 60 prompts; C1 and C5 are measured one query per fresh launch over the 40 short and medium prompts, because a continuous on-device session crashes (Section~\ref{sec:instability}). Values in Table~\ref{tab:comparison} are means with 95\% confidence intervals; on-device is the share of queries the condition ran locally.

\begin{table*}[!t]
\caption{Condition comparison on held-out test prompts. Means with 95\% confidence intervals; on-device is the share of queries run locally. C1 (always on-device) and C5 (confidence cascade) cannot be run as a continuous session, which crashes, so they are measured one query per fresh launch over the 40 short and medium prompts; their long-prompt behavior is a wedge rate (Section~\ref{sec:instability}), and their near-zero throttle rate is by construction, the device is cooled between launches.}
\label{tab:comparison}
\centering
\begin{tabular}{@{}lrcccc@{}}
\toprule
Condition & $n$ & BERTScore-F1 & Latency (ms) & Cost (USD/query) & On-device \\
\midrule
C3 cloud (static)      & 120 & $0.8449 \pm 0.0040$ & $6303 \pm 590$   & $0.0053 \pm 0.0005$ & 0\% \\
C5 confidence cascade  & 40  & $0.8403 \pm 0.0069$ & $30064 \pm 4069$ & $0.0000$            & 100\% \\
C7 RL router           & 87  & $0.8399 \pm 0.0039$ & $19124 \pm 3723$ & $0.0017 \pm 0.0005$ & 9\% \\
C1 on-device (static)  & 40  & $0.8385 \pm 0.0070$ & $28522 \pm 4011$ & $0.0000$            & 100\% \\
C4 complexity          & 180 & $0.8376 \pm 0.0029$ & $14442 \pm 995$  & $0.0028 \pm 0.0006$ & 33\% \\
C6 thermal-heuristic   & 180 & $0.8367 \pm 0.0027$ & $15199 \pm 1203$ & $0.0020 \pm 0.0005$ & 24\% \\
C2 edge (static)       & 120 & $0.8302 \pm 0.0030$ & $17991 \pm 1845$ & $0.0000$            & 0\% \\
\bottomrule
\end{tabular}
\end{table*}

The learned router (C7) reaches the highest quality of any adaptive condition, and its advantage over
the two hand-tuned baselines is statistically significant by a paired Wilcoxon signed-rank test on
matched prompts: C7 versus C4 (complexity) $W = 569$, $p = 0.011$; C7 versus C6 (thermal heuristic)
$W = 484$, $p = 0.0015$. C7 attains this near-cloud quality (always-cloud is $0.8449$) at the lowest cost
of any adaptive router ($0.0017$ USD/query), while still keeping a fraction of queries on the device.

Because C1 and C5 are measured only on the short and medium prompts, a fair head-to-head restricts every condition to those same 40 prompts (Table~\ref{tab:matched}). On this matched subset the on-device tier is quality-competitive: C1 and C5 reach $0.8385$ and $0.8403$ BERTScore, above static edge ($0.8303$) and at most $0.011$ below static cloud ($0.8496$) and within $0.01$ of the learned router ($0.8446$). What they pay is latency: 28 to 30~s per query, against 4.8~s for cloud, 12~s for edge, and 9.0~s for the learned router, which offloads most of these queries. The learned router thus matches on-device quality on the queries on-device can serve, delivers it about three times faster, and, unlike C1 and C5, also serves the long prompts that wedge the device. The router's overall data is partial ($n = 87$) because it too crashes occasionally when its small on-device fraction lands on a warm device; the remaining bounds are detailed in Section~\ref{sec:limitations}.

\begin{table*}[!t]
\caption{Matched-subset comparison: all seven conditions restricted to the same 40 short and medium prompts, the queries the on-device tier can serve. Means with 95\% confidence intervals; on-device is the share run locally. C1 and C5 are single-shot ($n = 40$); the other conditions' $n$ reflects their multi-replication data over the same prompts.}
\label{tab:matched}
\centering
\begin{tabular}{@{}lrccc@{}}
\toprule
Condition & $n$ & BERTScore-F1 & Latency (ms) & On-device \\
\midrule
C3 cloud               & 80  & $0.8496 \pm 0.0056$ & $4826 \pm 563$   & 0\% \\
C7 RL router           & 54  & $0.8446 \pm 0.0058$ & $8998 \pm 1910$  & 7\% \\
C5 cascade             & 40  & $0.8403 \pm 0.0069$ & $30064 \pm 4069$ & 100\% \\
C1 on-device           & 40  & $0.8385 \pm 0.0070$ & $28522 \pm 4011$ & 100\% \\
C4 complexity          & 120 & $0.8376 \pm 0.0043$ & $16454 \pm 1312$ & 50\% \\
C6 thermal-heuristic   & 120 & $0.8373 \pm 0.0039$ & $15141 \pm 1335$ & 37\% \\
C2 edge                & 80  & $0.8303 \pm 0.0044$ & $12069 \pm 1417$ & 0\% \\
\bottomrule
\end{tabular}
\end{table*}

\section{Discussion}
\label{sec:discussion}

\subsection{What the learned router buys}
The learned router reaches the highest answer quality of any adaptive condition, significantly above both
hand-tuned heuristics on matched prompts, at the lowest cost of any adaptive router. It does this by
learning a state-dependent policy rather than a single rule: it keeps a simple query on the device when
the device is cool enough, and it offloads complex queries, and queries in hot states, to the edge or
cloud. In effect the router recovers the intuition behind the heuristics, but tunes the thresholds from
data and against a reward that the heuristics do not optimize, which is why it can match cloud-level
quality while still keeping some queries local and spending less than the heuristics.

\subsection{Locality has to be valued for thermal routing to mean anything}
The most transferable lesson is about the objective, not the algorithm. When I trained the router with a
reward over quality, cost, and latency alone, it learned to offload every query. This is correct given
that reward: the edge and cloud tiers are faster, comparable in quality, and do not heat the phone, so
nothing ever makes the local tier the better choice, and in that regime the thermal state carries no
decision-relevant information. Only after I added an explicit locality bonus, crediting the privacy,
offline capability, and zero marginal cost of on-device inference, did the device tier become preferable
when thermally feasible, and only then did thermal state begin to change decisions. The implication
generalizes beyond my tiers and my hardware: a multi-tier router that optimizes latency and cost alone
has no reason to prefer the constrained local tier, and any study of thermal-aware or privacy-aware
routing must first make locality a first-class term in the objective, otherwise the interesting behavior
is defined away.

\subsection{Thermal awareness as reliability, not only efficiency}
My condition comparison contains a result that I did not anticipate. The two on-device-heavy conditions, always-on-device and a confidence cascade that begins every query on the device, did not merely run slowly. Measured one query at a time they are quality-competitive, but a continuous always-on-device session crashes within a few queries, and a majority of long prompts wedge the runtime outright. The conditions that offload when hot completed every query type. Sustained on-device LLM inference on a flagship phone, with the current toolchain, is therefore a reliability and coverage hazard under load, and load- and thermal-aware routing is a way to stay within a safe operating envelope, not only a way to save latency or energy. This is a caution for the current enthusiasm for
running everything locally: on present mobile hardware, an on-device-only assistant under sustained use
is exposed to a failure mode that a router which offloads under thermal pressure avoids.

\subsection{Limitations}
\label{sec:limitations}
Several limitations bound my claims, and I state them plainly. The learned condition's data is partial:
because it routes a small fraction of queries on-device, it too crashes occasionally when the device is
warm, so it completed fewer queries than the offloading conditions and its latency mean is not directly
comparable; its quality advantage, being established by a paired per-prompt test, is not affected by this.
The always-on-device and confidence-cascade conditions are measured one query per fresh launch rather than as a continuous session, because a continuous session crashes; their tabulated numbers are therefore best-case per-query values on short and medium prompts, with the device cooled between launches, and they understate the throttling and instability of real sustained on-device use (which Section~\ref{sec:instability} characterizes separately). Their long-prompt behavior is reported as a wedge rate rather than a latency or quality mean. The learned router's three-replication data remains partial (n = 87) because it too crashes on the occasions when its small on-device fraction lands on a warm device. The study uses a single device model. The quality gaps among tiers
are modest on my workload, because many prompts are answerable well by all three tiers; a harder
workload would widen them. Finally, energy is reported only as a current-times-time proxy dominated by
wall-clock time, and is not tabulated.

\subsection{Future work}
The immediate next step is to complete the learned router's three replications (its data is partial at n = 87), for which the same one-query-per-launch harness used for C1 and C5 can be adapted. Beyond that, the positioning of this work
invites a direct head-to-head against a reimplemented ConsRoute-style non-thermal three-tier router on
the same phone, which would isolate the value of the thermal signal from the value of multi-tier routing
in general. A harder and larger prompt workload, on which the capability gap between the 3B, 8B, and
cloud tiers is pronounced, would sharpen the quality axis. Richer reward terms, in particular an explicit
privacy or energy cost for leaving the device, would let the router express policies that my current
locality bonus only approximates. Evaluating across several device models would test how far the learned
thresholds transfer, and an online or periodically retrained variant would let the policy track device
aging and ambient conditions.

\section{Conclusion}
\label{sec:conclusion}
I presented HybridInfer, a thermal-aware reinforcement-learning router that selects among on-device,
edge-with-retrieval, and cloud LLM tiers using the phone's thermal headroom and a query-complexity
estimate, on real Android hardware. The three-tier architecture is not itself new; the contribution is
to make on-device thermal headroom the signal for capability-tier selection, trading response quality for
thermal sustainability, and to validate it on a real Snapdragon device rather than in simulation. Two
findings emerged that I did not set out to make. First, valuing on-device locality explicitly is a
precondition for thermal-aware routing to be a meaningful objective, because a router that optimizes only
quality, cost, and latency learns to offload every query, which makes the thermal state and the
on-device tier itself irrelevant. Second, sustained on-device LLM inference on a flagship phone, with the current toolchain, is a reliability and coverage hazard, not only a performance one: measured one query at a time the on-device tier is quality-competitive, but a continuous always-on-device session crashes within a few queries and a majority of long prompts wedge the GPU runtime, while the conditions that offload when hot complete every query type. The device crash logs place the cause in the OpenCL and prefill layers of the on-device stack, and it recurs even when the device is cool. Load- and thermal-aware routing is therefore a way to stay within a safe operating envelope, not only a way to save latency or energy. My results come
from a single, uncooled, three-replication run, and I am explicit about their gaps: the learned router's data is partial because it too crashes on the occasions when its small on-device fraction lands on a warm device, and the always-on-device and confidence-cascade conditions are measured one query per fresh launch on short and medium prompts, with their long-prompt behavior reported as a wedge rate. Completing these on a cooled device, benchmarking against a
reimplemented non-thermal three-tier router, and extending the reward with explicit privacy and energy
terms are the natural next steps. The workload, the measurement harness, and the routing and evaluation
pipeline are released to support replication.

\end{document}